\documentclass[twoside,leqno,twocolumn]{article}

\usepackage[letterpaper]{geometry}

\usepackage{array}
\newcolumntype{P}[1]{>{\centering\arraybackslash}p{#1}}

\usepackage{ltexpprt}
\usepackage{hyperref}

\usepackage{graphicx}
\usepackage{amsmath, bm}
\usepackage{amsfonts}
\usepackage{float}
\usepackage{varioref}
\usepackage{multirow}
\usepackage{multicol}
\usepackage{subcaption}
\usepackage{algorithm}
\usepackage{algorithmic}

\usepackage{tabularx}
\begin{document}



\title{\Large Cycle-Balanced Representation Learning For Counterfactual Inference
}
\author{
Guanglin Zhou
\thanks{University of New South Wales. \newline
Email: \{guanglin.zhou, lina.yao\}@unsw.edu.au}
\and Lina Yao\footnotemark[1]

\and Xiwei Xu\thanks{Data61, CSIRO. \newline
Email:\{xiwei.xu,  chen.wang, liming.zhu\}@data61.csiro.au}
\and Chen Wang \footnotemark[2]
\and Liming Zhu \footnotemark[2]

}

\date{}

\maketitle







\begin{abstract} \small\baselineskip=9pt 
With the widespread accumulation of observational data, researchers obtain a new direction to learn counterfactual effects in many domains (e.g., health care and computational advertising) without Randomized Controlled Trials (RCTs). However, observational data suffer from inherent missing counterfactual outcomes, and distribution discrepancy between treatment and control groups due to behaviour preference. Motivated by recent advances of representation learning in the field of domain adaptation, we propose a novel framework based on \textsc{\bf C}ycle-\textsc{\bf B}alanced \textsc{\bf RE}presentation learning for counterfactual inference (CBRE), to solve above problems. Specifically, we realize a robust balanced representation for different groups using adversarial training, and meanwhile construct an information loop, such that preserve original data properties cyclically, which reduces information loss when transforming data into latent representation space. Experimental results on three real-world datasets demonstrate that CBRE matches/outperforms the state-of-the-art methods, and it has a great potential to be applied to counterfactual inference.
\end{abstract}

\section{Introduction.}
Inferring counterfactual effects is a fundamental problem in many applications like health care \cite{mani2000causal, alaa2017bayesian, glass2013causal}, economics \cite{chernozhukov2013inference, li2016matching},  computational advertising \cite{JMLR:v14:bottou13a} and education \cite{zhao2017estimating}. This can be defined as a what-if question, such as, teachers want to know whether the grades of  students would improve if they had selected another teaching method. The gold standard for estimating the effects are Randomized Controlled Trials (RCTs) \cite{autier2007vitamin}. In RCTs, the treatment assignment is controlled by researchers and it makes sure that the assignment approach is independent to users or patients. Researchers obtain an unbiased estimator by performing RCTs, due to the randomness of assigning users or patients to either the treatment or control group. However, acquisition of RCTs data is expensive, time-consuming, and sometimes immoral. For example, in the field of online advertising \cite{JMLR:v14:bottou13a}, algorithm designers perform RCTs and divide website traffic into different groups, in pursuit to evaluate the performance of recommendation algorithms fairly and objectively. The evaluation period may be several months and damage user experience if the algorithm performance is poor.      

In contrast to RCTs data, observational data, that exist in what are known as observational studies, normally consist of a record of the input, an intervention and the corresponding outcome. For example, the electronic medical records for diabetics are typical observational data. There are some statistical information about the patients in the records like height, weight, gender, age, income and job. We call them co-variables related to the input. The treatment approach arranged by doctors, such as medicine treatment or surgery treatment, and the final treatment effect that the patient is cured or not cured half a year later are also recorded. This type of data are widely available and comparatively easy to acquire, which has been playing an increasing role in estimating counterfactual effects.

There are two major challenges that remain in this area. One is the problem of distribution discrepancy between the treatment and control group due to behaviour preference. Different from data in RCTs, there is no random setting in observational data and therefore the problem of bias arises, such as selection bias \cite{chen2020bias}. For example, when selecting a treatment for diabetes, patients with worse socioeconomic status are used to choosing relative cheaper treatment approaches rather than more advanced and meanwhile expensive treatment solutions. User preference brings distribution shift between the treatment and control group. This is why we cannot solve the problem by using supervised learning simply. There is no guarantee of independently and identically distributed $(i.i.d)$ for either training data or test data. The second challenge is that we miss the counterfactual outcomes. A unit, such as one diabetic, always belongs to one group, in other words, the patient can only choose one treatment for his disease at any specific time. Therefore, the outcome of another treatment (called counterfactual) is always missing.

\begin{figure}[tbp!]
    \centering
    \includegraphics[width=\linewidth]{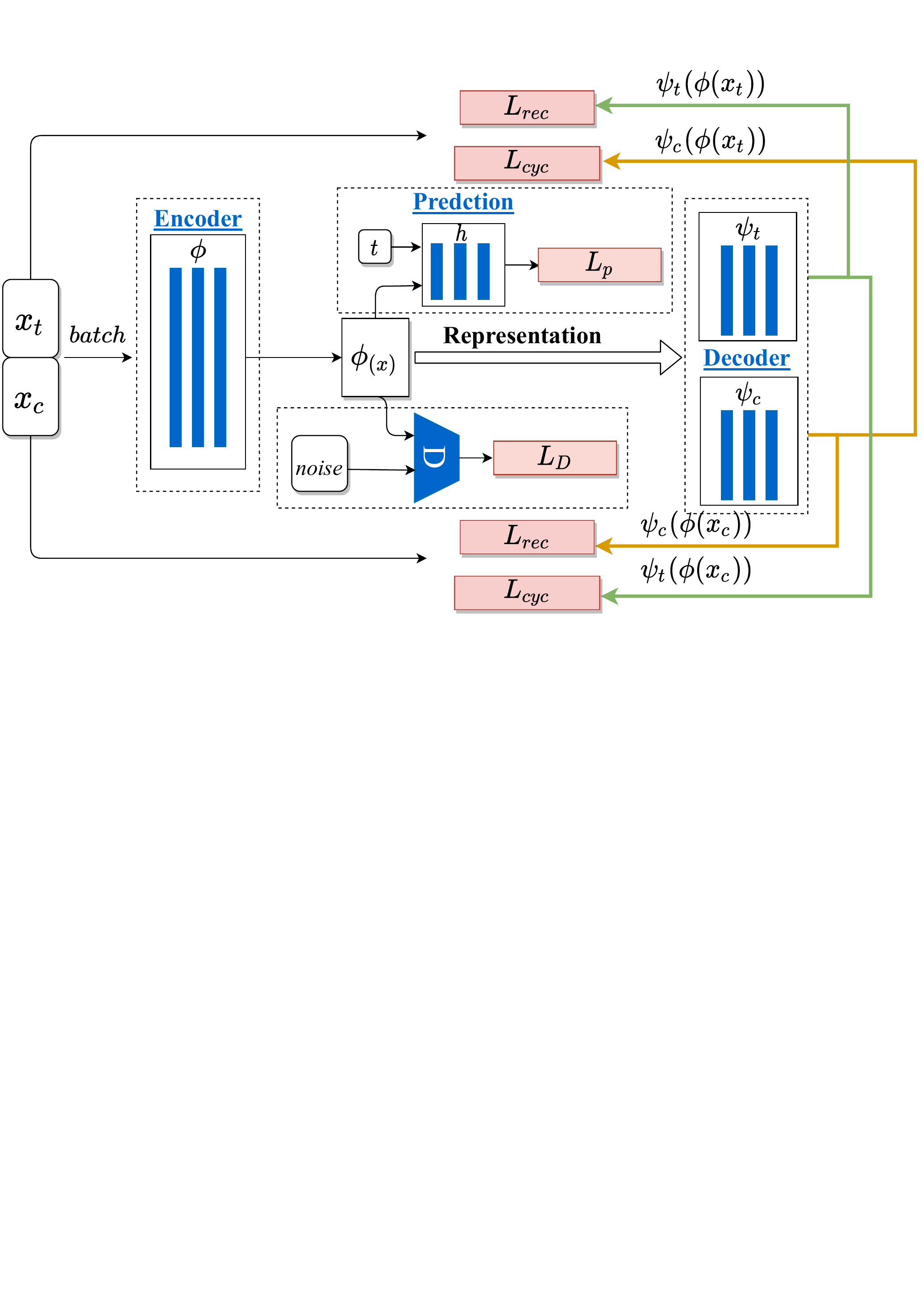}
    \vspace*{-6.5cm}
    \caption{The overview of our model. Data from different groups are fed into the model by batch. Blue denotes components of the model: one encoder, one discriminator, two decoders and one prediction network. Pink expresses loss functions: the adversarial loss $\mathcal{L}_{D}$, the prediction loss $\mathcal{L}_p$, the information loss $\mathcal{L}_{rec}$ and $\mathcal{L}_{cyc}$. The information loop is indicated by yellow and green lines.} 
    \label{fig:CBRE_model}
\end{figure}

In this work, we introduce an information loop (as Figure \ref{fig:CBRE_model} shows) to construct an unbiased estimator for individual treatment effect (ITE) from observational data. Our model balances representations of two groups and meanwhile further preserves salient information cyclically for better prediction effects. We reduce the distribution shift of latent representations between the treatment and control group in an adversarial training manner. Then, we decode representations to raw data space, and define information reconstruction and cycle loss to preserve highly predictive information for outcome prediction network.


The key contributions are provided as follows. 
\begin{itemize}
    \item We propose a novel framework based on cycle-balanced representation learning to infer counterfactual effects, by considering both the distribution discrepancy and preserving original data properties cyclically.
    \item We encode raw data from both treatment and control groups into latent representation space to generate latent features. We align the distribution of latent vectors by using an adversarial training method. When the discriminator cannot distinguish which group the latent feature vector comes from, we eliminate negative effects from distribution shift.
    \item As highly predictive information may be lost when transforming raw data into latent space, we design decoder networks to enable information reconstruction and compensate for information loss in transformation cycle. 
    \item We perform experiments on three real-world datasets with 10 baseline models. And the results demonstrate the effectiveness and generalization ability of our model.
\end{itemize}


\section{Problem Setup.}
We employ the following notations and assumptions in our work. The summary of notations can be found in supplementary material. We consider a setting in which we are given observational dataset $\mathcal{D}=\{x_i, t_i, y_i\}_{i=1}^{n}$ from an existing but unknown joint distribution, where $x_i \in \mathcal{X} \subseteq {\mathbb{R}}^{p}$  represents covariates matrix related to patients. The treatment $t_i$ is selected from a set $\mathcal{T}$ (e.g., \{0: medicine treatment, 1: surgery treatment\}) and $y_i \in \mathcal{Y} \subseteq \mathbb{R}$ denotes observed outcome (e.g., binary outcome: \{0: not cured, 1: cured\} or continuous outcome: blood sugar).

For unit $i$, we get $p$ covariates, $x_i^{(0)},x_i^{(1)},...,x_i^{(p-1)}$, associated with treatment assignment $t_i$ and observed outcome $y_i$. In this work, we focus on binary treatment effects with treatment $t_i\in\{0, 1\}$. Specifically, we define raw data in treatment group as $x_t$ with treatment set $\mathcal{T}=1$, and express raw data in control group as $x_c$ with treatment set $\mathcal{T}=0$. We follow the potential outcome framework proposed by Rubin-Neyman \cite{rubin1974estimating, rubin2005causal, rubin2001using}. Therefore, there are two potential outcomes for each unit $i$, 
$y_i^{t_i}$ and $y_i^{1-t_i}$.
In fact, we only observe one outcome in observational data. We denote $y_i^{t_i}$ as factual outcome $y_i^{F}$ and $y_{i}^{1-t_i}$ as counterfactual outcome $y_i^{CF}$. Our aim is to precisely estimate $y_i^F$ and $y_i^{CF}$ for each unit $i$, in spite of distribution discrepancy between the treatment and control group. 


            



Following the potential outcome framework \cite{rosenbaum1983central}, we make one definition and some common assumptions in our work.  

\textit{Definition 2.1.} \textbf{Individual Treatment Effect (ITE).} The individual treatment effect means that, for each individual unit $i$, the potential outcome difference between treatment and control group:
\begin{equation} \label{eq:1}
    ITE_{i}=y_{i}^{t_i=1}-y_{i}^{t_i=0}
\end{equation}

\textit{Assumption 2.1}. \textbf{Unconfoundedness.} Confounders represent some covariates that are both used to assign treatments and related to the outcome. In RCTs, $\mathcal{X}$ is known to include all covariates. In nonrandomized experiment like observational study, we make the assumption of unconfoundedness. Given the variable $x_i$, the outcome $y_i$ is conditionally independent on treatment assignment $t_i$, i.e., $y_i^{t_i} \perp\!\!\!\!\perp t_i | x_i $.         

\textit{Assumption 2.2}. \textbf{Overlap.} Same with RCTs that every individual in the population has a chance of receiving each treatment, we make the overlap assumption. Formally, $0<P_r(\mathcal{T}=t|\mathcal{X}=x)<1,\  \forall x \in \mathcal{X},\forall t \in \mathcal{T} $.  

\textit{Strong Ignorability} consists of the above two assumptions. Within the strong ignorability assumption, we are able to tackle the problem of approximating potential outcome using a machine learning model $f: \mathcal{X} \times \mathcal{T} \rightarrow$ $\mathcal{Y}$.  

In this work, we determine the model $f$ from two aspects. We not only consider the distribution shift between two groups, but also further explore inherent information loss and propose a paradigm of the information loop.

\section{The Proposed Method.} 
In the basis of above problem setup, we propose a cycle-balanced representation learning method for counterfactual inference, aiming to eliminate the distribution discrepancy and preserve highly predictive information.  


As the Figure \ref{fig:CBRE_model} shows, there are four parts in our model: an encoder function $\phi: \mathcal{X}\rightarrow \mathcal{Z}$ that maps the raw covariates space $\mathcal{X}$ into latent representation space $\mathcal{Z}$, two decoder functions corresponding to the treatment and control group $\psi_t:\mathcal{Z}\rightarrow \mathcal{X}$ and $\psi_c:\mathcal{Z}\rightarrow \mathcal{X}$, a discriminator $f_D$ that is used to distinguish whether $\phi_{(x_i)}$ comes from the treatment or control group, a prediction network $h: \mathcal{Z} \times \mathcal{T}\rightarrow \mathcal{Y}$ that predicts factual and counterfactual outcomes given $\mathcal{X}$ and $\mathcal{T}$.  

Accordingly, we define loss functions in our model. The loss in the discriminator is expressed as $\mathcal{L}_D$ to measure the balance extent between the treatment and control group in latent space $\mathcal{Z}$. The information reconstruction loss, $\mathcal{L}_{rec}$, trains two separate decoders to reconstruct $\mathcal{X}$ from latent representations $\phi_{(\mathcal{X})}$. And cycle loss $\mathcal{L}_{cyc}$ is used to further retain highly predictive information. Finally batches of representations $\phi_{(x_t)}$ and $\phi_{(x_c)}$ are fed forward to outcome prediction network, which generates factual loss $\mathcal{L}_{p}$ between the estimated and observed factual outcomes.

The total loss is sum of above losses and defined as follows:
\begin{equation} \label{eq:2}
    \mathcal{L} = \mathcal{L}_{p} + \alpha \mathcal{L}_{D} + \beta \mathcal{L}_{rec} + \gamma \mathcal{L}_{cyc} + \lambda \left \| W \right \|_2 
\end{equation}
The final term $\left \| \cdot \right \|_2 $ is $l_2$ regularization for model complexity \cite{ng2004feature}. We use $W$ to denote the model parameters, where $\lambda \geq 0$ means the trade-off between the $l_2$ regularization and other losses.

\subsection{Balance Distribution Discrepancy}

Due to the characteristic of observational data, distribution discrepancy exists between the treatment and control group. To reduce distribution shift, we aim at learning representations of raw data and making latent representations from two groups similar. We learn an encoder $\phi: \mathcal{X} \rightarrow \mathcal{Z}$ and a discriminator $f_D$, aiming to eliminate the distribution discrepancy. 
Specifically, we transform original features $x_t$ from the treatment group into latent representation space, which is named as $\phi_{(x_t)}$. And correspondingly, we get $\phi_{(x_c)}$ and align two representations as similar as possible.
The discriminator $f_D$ is used to measure and control the similarity. We assume that we eliminate the distribution shift in the latent space if the discriminator cannot distinguish which group the latent representations come from.  

As the Figure \ref{fig:CBRE_model} shows, we take latent representations and a noise vector as inputs of the discriminator. More specifically, we derive the noise vector from a Gaussian distribution $\boldsymbol{v} \sim \mathcal{N}(0,1)$, which has same dimension with the representations. In the latent space, we divide the representations by treatment assignment into the treatment group $z=\phi_{(x_t)}, z \sim \mathbb{P}_t$ and control group $z=\phi_{(x_c)}, z \sim \mathbb{P}_c$. Then we concatenate noise vector with two group of representations separately, and we get $[\boldsymbol{v}, z], z \sim \mathbb{P}_t$ along with $[\boldsymbol{v}, z], z \sim \mathbb{P}_c$. We play a minimax game between two competing networks of the encoder and discriminator. The encoder is trained to fool the discriminator.

Formally, the minimax objective is:
\begin{equation} \label{eq:3}
    \mathop{min}\limits_{\phi}\mathop{max}\limits_{f_D} \mathop{\mathbb{E}}_{z\sim \mathbb{P}_t}[log(f_D([\boldsymbol{v},z]))] + 
    \mathop{\mathbb{E}}_{ z \sim \mathbb{P}_c}[log(1-f_D( [\boldsymbol{v},z]))]
\end{equation}

Considering the instability of training in Generative Adversarial Networks (GANs) \cite{goodfellow2014generative}, we adopt Wasserstein GANs to achieve stable training and better performance \cite{10.5555/3295222.3295327}. WGAN utilizes 1-Lipschitz functions to minimize the Earth-Mover distance $W(\mathbb{P}_t, \mathbb{P}_c)$ between the distributions of treatment and control group in embedding space, with respect to the generator parameters. 
\begin{equation} \label{eq:wgan}
    \mathop{min}\limits_{\phi}\mathop{max}\limits_{f_D} \mathop{\mathbb{E}}_{z\sim \mathbb{P}_t}[f_D([\boldsymbol{v},z])] - 
    \mathop{\mathbb{E}}_{ z \sim \mathbb{P}_c}[f_D( [\boldsymbol{v},z])]
\end{equation}
$f_D$ should be the set  of 1-Lipschitz functions. We also add a penalty on the gradient norm to enforce the 1-Lipschitz constraint. We define $\mathbb{P}_{penalty}$ to sample uniformly pairs from the data distribution $\mathbb{P}_t$ and $\mathbb{P}_c$. 

Thus, we get the final loss function defined as:
\begin{equation} \label{eq:4}
\begin{aligned}
    \mathcal{L}_D=\mathop{\mathbb{E}}_{z \sim \mathbb{P}_c}[f_D([\boldsymbol{v},z])] - \mathop{\mathbb{E}}_{z \sim \mathbb{P}_t}[f_D([\boldsymbol{v},z])] + \\ \delta\cdot\mathop{\mathbb{E}}_{z \sim \mathbb{P}_{penalty}}[(\left \| \bigtriangledown_{[\boldsymbol{v},z]} f_D([\boldsymbol{v},z]) \right \|_2-1)^2]
\end{aligned}
\end{equation}
By minimizing $\mathcal{L}_D$, we make sure that the encoder $\phi$ generates balanced representations of two groups in order to fool the discriminator. $\phi$ and $f_D$ can be optimized by stochastic gradient decent.

\subsection{Cycle-Preserve Information}
Using above encoder and discriminator modules, we are able to measure and control the distribution discrepancy. However, it causes an inherent problem that information loss exists when transforming raw data into latent space. It indeed makes sure distribution discrepancy can be balanced but it is unavoidable to miss highly predictable information. This part of information actually represents user's preference and selections \cite{chen2020bias}. We start with the information loop. Specifically, we add two separate decoders (see $\psi_{t}$ and $\psi_{c}$ in Figure \ref{fig:CBRE_model}) corresponding to the treatment and control group, and define information reconstruction and cycle loss. The decoder modules constrain the encoder and preserve original data properties for better outcome estimation.

Firstly, we reconstruct input raw covariables by using the corresponding decoder. For example, we get the latent representations $\phi_{(x_t)}$ from batch of $x_t$ and reconstruct $x^{'}_t: \psi_{t}{(\phi{(x_t)})}$ using the decoder $\psi_t$. Similarly, we get reconstruction $x^{'}_c: \psi_{c}{(\phi{(x_c)})}$.  

we use $\mathcal{L}_{rec}$ to measure the reconstruction loss from latent space into raw data space:
\begin{equation} \label{eq:5}
\begin{aligned}
    \mathcal{L}_{rec} = \frac{1}{N_t}\sum_{n=1}^{N_t} \left \|  x_t-\psi_t({\phi{(x_t)}})\right \|^2 + \\ \frac{1}{N_c}\sum_{n=1}^{N_c} \left \|  x_c-\psi_c({\phi{(x_c)}})\right \|^2
\end{aligned}
\end{equation}

In order to encourage the informative contents to be preserved and inspired by domain adaptation \cite{hoffman2018cycada, chen2019distributionally}, we add a cycle-consistency constraint on decoders. Intuitively, we treat the treatment distribution as target domain and the control distribution as source domain. The cycle-consistency means that we map a sample from the source domain to the target domain, and then map it back to the source domain, which needs to be consistent with the original sample. This can be formalized as $M_{T \rightarrow S}(M_{S \rightarrow T}(x_s)) \approx x_s$ where $M$ is a set of mapping functions. In the context of treatment and control distributions, we utilize the encoder and decoders as the mapping functions, and expect $\psi_t(\phi(\psi_c(\phi(x_t))))$ to be approximately equal to $x_t$. It can be simplified as $\psi_c(\phi(x_t)) \approx x_t$, and similarly for the reverse: $\psi_t(\phi(x_c)) \approx x_c$. 

Therefore, information preserving is further conducted by the cycle loss $\mathcal{L}_{cyc}$ in the Figure \ref{fig:CBRE_model}:
\begin{equation} \label{eq:6}
\begin{aligned}
    \mathcal{L}_{cyc} = \frac{1}{N_t}\sum_{n=1}^{N_t} \left \|  x_t-\psi_c({\phi{(x_t)}})\right \|^2 + \\ \frac{1}{N_c}\sum_{n=1}^{N_c} \left \|  x_c-\psi_t({\phi{(x_c)}})\right \|^2
\end{aligned}
\end{equation}  

In both Eq.\eqref{eq:5} and \eqref{eq:6}, $N_t$ and $N_c$ refer to batch numbers for $x_t$ and $x_c$. $\left \| \cdot \right \|^2$ refers to a distance metric.   

By the aid of $\mathcal{L}_{rec}$ and $\mathcal{L}_{cyc}$, we build the information loop that constrains the encoder module and preserves original data properties. 

\subsection{Outcome Prediction Network}  

With the above two components, the model is able to balance the distribution shift and preserve highly-predictive information cyclically. And we employ the outcome prediction network, that takes the representation $\phi(x_i)$ and treatment assignment $t_i$ as inputs. As the Figure \ref{fig:CBRE_model} shows, we denote $h(\cdot)$ as the function learned by the outcome prediction network. We use mean square error between predicted outcomes and observed factual outcomes to express the loss $\mathcal{L}_{p}$. And in purpose of better performance and fair comparison, we employ the weights used in  \cite{shalit2017estimating}.  
So the loss is as follows:
\begin{equation} \label{eq:7}
    \mathcal{L}_p = \frac{1}{n}\sum_{i=1}^n \omega_i \cdot (y_{i} - h(\phi_{(x_i)}, t_i))^2
\end{equation}
with $\omega_i=\frac{t_i}{2u}+\frac{1-t_i}{2(1-u)}$, where $u=\frac{1}{n}\sum_{i=1}^{n}t_i$, and $n$ is the number of units.  

Once we have the encoder $\phi_{(x_i)}$ and prediction network $h(\cdot)$  with good performance, we can estimate factual and counterfactual outcomes: $\hat{y}_{i}^{F}=h(\phi_{(x_i)}, t_i)$ and $\hat{y}_{i}^{CF}=h(\phi_{(x_i)}, 1-t_i)$

\subsection{Training and Optimization}
The training objective is to minimize total loss function of Eq.\eqref{eq:2}. We use fully-connected feed-forward neural networks to model the encoder, discriminator, two decoders and prediction network separately. We use Dropout 
and Relu 
as activation functions. Adam 
, that is a stochastic gradient-based optimization method, is used to optimize the loss function of Eq.\eqref{eq:5} jointly at a learning rate of $1e$-$3$. Batch Normalization is also used to get better performance. The detailed procedure is shown in 
Supplementary material.

\section{Experiments}
In this section, we evaluate the performance of our method on three real-world datasets. Firstly, we describe the datasets, baselines we compare to, and the metrics for evaluating the performance. We compare to state-of-the-art models that cover all three categories in \nameref{related works}. Secondly, we introduce and analyse results on three datasets. Then, a detailed ablation study is conducted to examine each component in CBRE for inference performance. And we perform the t-SNE visualization of raw data and corresponding latent representations. The visualization proves the effectiveness for balancing distribution shift. At last, we introduce optimal hyper-parameters.  

In this work, we perform all experiments on a cluster with two 12-core Intel Xeon E5-2697 v2 CPUs and a total 768 GiB Memory RAM.

\subsection{Dataset Description}

It's difficult to evaluate counterfactual inference models, due to the lack of the ground truth treatment effect. In other words, we have no access to counterfactual outcomes in real world. In this paper, we adopt two common methods like synthetic and semi-synthetic, which are widely used in state-of-the-art models.

 We use two semi-synthetic datasets, IHDP and Twins, where either treatment assignments or potential outcomes are synthesized. And one real-world dataset Jobs that combines Randomized Controlled Trials (RCTs) and observational study.  
 The summary of datasets is shown as Table \ref{table_datasets}. Details about the three datasets are provided in the supplementary material.
 
\begin{center}
\begin{table}[h!]
\caption{The summary of three datasets. RCTs means whether the data source contains data from RCTs.}
\label{table_datasets}
\centering
\begin{tabular}{|c|c|c|c|}
\hline
              \textbf{Property} & \textbf{IHDP} & \textbf{Jobs} & \textbf{Twins}  \\ \hline
              {F} & {$\checkmark$} & {$\checkmark$} & {$\checkmark$}  \\ \hline
              {CF} & {$\checkmark$} & {$\times$} & {$\checkmark$}  \\ \hline
              {Treatment} & {Binary} & {Binary} & {Binary}  \\ \hline
              {RCTs} & {$\times$} & {$\checkmark$} & {$\times$}  \\ \hline
              {Num} & ${747}$ & ${3212}$ & ${11400}$  \\ \hline
              {Dimension} & ${25}$ & ${8}$ & ${30}$  \\ 

\hline            

\end{tabular}
\end{table} 
\end{center}

\subsection{Experiment Setting}
We describe the baseline methods which represent state-of-the-art models for counterfactual inference.  

\textbf{Baselines.} We compare the proposed method with the following 10 baselines: least square regression using treatment as a feature \textbf{(OLS/LR}$_1$\textbf{)};
separate least square regressions for each treatment group \textbf{(OLS/LR}$_2$\textbf{)}; 
Bayesian additive regression trees \textbf{(BART)}  \cite{chipman2010bart}; 
k-nearest neighbor \textbf{(k-NN)}  \cite{crump2008nonparametric}; 
balancing linear regression \textbf{(BLR)} and balancing neural networks \textbf{(BNN)} that firstly connects representation learning with counterfactual inference  \cite{johansson2016learning}; 
treatment-agnostic representation networks \textbf{(TARNet)} and counterfactual regression with Wasserstein distance \textbf{(CFR-Wass)}  \cite{shalit2017estimating}; 
local similarity preserved individual treatment effect \textbf{(SITE)} \cite{yao2018representation}; 
adversarial balance for causal effect inference \textbf{ABCEI} \cite{du2021adversarial}.  

\textbf{Metrics.} On IHDP dataset, the Rooted Precision in Estimation of Heterogeneous Effect ($\sqrt{\epsilon_{PEHE}}$) and Mean Absolute Error on ATE ($\epsilon_{ATE}$) are used as performance metrics. The smaller of two metrics are, the better the performance is. Formally, the definitions are:
\begin{equation} \label{eq:8}
    \sqrt{\epsilon_{PEHE}} = \sqrt{\frac{1}{n}\sum_{i=1}^n ((y_i^{t=1}-y_i^{t=0})-(\hat{y}_i^{t=1}-\hat{y}_i^{t=0}))^2}
\end{equation}
\begin{equation} \label{eq:9}
    \epsilon_{ATE} = \left| \frac{1}{n}\sum_{i=1}^n (y_i^{t=1}-y_i^{t=0})-\frac{1}{n}\sum_{i=1}^n(\hat{y}_i^{t=1}-\hat{y}_i^{t=0}) \right|
\end{equation}

where $y_i$ denotes the observed outcome and $\hat{y_i}$ denotes the predicted outcome.  
Following \cite{shalit2017estimating}, we use policy risk estimation $R_{pol}$ to measure performance for Jobs dataset. We seek smaller $R_{pol}$ on Jobs.
\begin{equation} \label{eq:10}
\begin{aligned}
    R_{pol}(\pi) = 1- [\mathbb{E}(y_i^{t_i=1}|\pi(x_i)=1)\cdot P(\pi(x_i)=1) \\
    +\mathbb{E}(y_i^{t_i=0}|\pi(x_i)=0)\cdot P(\pi(x_i)=0)]
\end{aligned}
\end{equation}
where $\pi(x_i)=1$ if $\hat{y_i}^{t_i=1}-\hat{y_i}^{t_i=0}>0$, and $\pi(x_i)=0$, otherwise.

\begin{tiny}
\begin{table*}[!htb]
\caption{Performance Evaluation of $\textbf{CBRE}$ with other state-of-the-art methods on three datasets of IHDP, Jobs and Twins. Bold indicates the method with the best performance.}
\label{table-performance}
\centering
\begin{tabularx}{\textwidth}{X|X|X|X|X|X|X|X|X}
\hline
\multirow{3}{*}{Methods}   & \multicolumn{4}{c}{\textbf{IHDP}}    & \multicolumn{2}{c}{\textbf{Jobs}} & \multicolumn{2}{c}{\textbf{Twins}}                                                     \\ \cline{2-9}
                           & \multicolumn{2}{c}{In-sample} & \multicolumn{2}{c}{Out-sample} & \multicolumn{1}{c}{In-sample} &  \multicolumn{1}{c}{Out-sample}  & \multicolumn{1}{c}{In-sample} & \multicolumn{1}{c}{Out-sample}               \\ \cline{2-9}
                           & $\sqrt{\epsilon_{PEHE}}$ & $\epsilon_{ATE}$ & $\sqrt{\epsilon_{PEHE}}$ & $\epsilon_{ATE}$  & $R_{pol}$  & $R_{pol}$ & $AUC$ & $AUC$\\ \hline
            OLS/LR$_1$     & $5.8\pm .3$                & $.73\pm .04$   & $5.8\pm .3$               & $.94\pm .06$  & $.22 \pm .0$            & $\textbf{.23}\pm \textbf{.0}$ &  $.66 \pm .01$                & $.50\pm .03$    \\
            OLS/LR$_2$     & $2.4\pm .1$                & $.14\pm .01$   & $2.5\pm .1$               & $.31\pm .02$  & $.21\pm .0$             & ${.24\pm .0}$ & $.66\pm .00$                & $.50\pm .02$    \\
            BLR     & $5.8\pm .3$                & $.72\pm .04$   & $5.8\pm .3$               & $.93\pm .05$  & $.22\pm.0$                 & ${.25\pm.0}$  & $.61\pm.01$                & $.51\pm .02$   \\
            BART     & $2.1\pm .1$                & $.23\pm .01$   & $2.3\pm .1$               & $.34\pm .02$  & $.23\pm.0$                & ${.25\pm.0}$   & $.51\pm.01$                & $.50\pm.01$  \\
            k-NN         & $2.1\pm .1$                & $.14\pm .01$   & $4.1\pm .2$               & $.79\pm .05$   & $.23\pm.0$                   & ${.26\pm.0}$   & $.61\pm.01$                & $.49\pm.01$   \\ \hline
            BNN     & $2.2\pm .1$                & $.37\pm .03$   & $2.1\pm .1$               & $.42\pm .03$   & $.20\pm.0$                 & ${.24\pm.0}$   & $.69\pm.01$                & $.68\pm.01$ \\
            TARNet     & $.90\pm .0$                & $.25\pm .03$   & $.98\pm .1$               & $.27\pm .04$     & $.17\pm.0$                   & ${.32\pm.0}$  & $.48\pm.00$                & $.48\pm.01$   \\
            CFR-Wass     & $.77\pm .0$                & $.29\pm .04$   & $.83\pm .1$               & $.30\pm .04$   & $.16\pm.0$                   & ${.30\pm.0}$   & $.84\pm.00$                & $.86\pm.01$    \\
            SITE         & $.91\pm .0$                & $.27\pm .04$   & $.95\pm .1$               & $.27\pm .03$  & $.16\pm.0$                   & ${.29\pm.0}$  & $.76\pm.00$                & $.78\pm.01$      \\
            ABCEI        & $.78\pm .1$                & ${.10}\pm {.01}$   & $.92\pm .1$               & ${.14}\pm {.02}$ & $.16\pm.0$                   & ${.37\pm.0}$  & $.86\pm .00$                & $.88\pm.00$     \\ \hline
            CBRE         & $\textbf{.52}\pm \textbf{.0}$       & $\textbf{.10}\pm \textbf{.01}$   & $\textbf{.60}\pm \textbf{.1}$               & $\textbf{.13}\pm \textbf{.02}$   & $\textbf{.13}\pm \textbf{.0}$       & ${.28\pm.0}$   & $\textbf{.86}\pm \textbf{.00}$  & $\textbf{.88}\pm\textbf{.00}$    \\
\hline

\end{tabularx}
\end{table*} 
\end{tiny}

As suggested in \cite{yao2018representation}, we adopt area over ROC curve (AUC) on outcomes as the performance measure for Twins dataset. If AUC is larger, the performance is better.

\subsection{Results}

Results in Table \ref{table-performance} demonstrate effectiveness of our method compared with state-of-the-art models. we perform 100 experiments and report mean results on all of three datasets. And we split training/validation/test sets with 60\%/30\%/10\%. Generally speaking, our method achieves best performance on IHDP and Twins, and achieves similar results or outperforms state-of-the-art models on Jobs. We achieve seven best results among eight indicators.

\begin{table*}[!ht]
\caption{Ablation Study. $\mathcal{L}_p$ means an encoder plus prediction network; $\mathcal{L}_p+\mathcal{L}_d$ denotes the model with extra reduction in distribution shift; $\mathcal{L}_p+\mathcal{L}_{rec}+\mathcal{L}_{cyc}$ is the model that cares information preservation; Total model is our proposed model.}
\label{table_Ablation Study}
\centering
\begin{tabular}{cccccc}
\hline
              Dataset    &    & Total model &  $\mathcal{L}_p$ &            $\mathcal{L}_p+\mathcal{L}_d$     & $\mathcal{L}_p+\mathcal{L}_{rec}+\mathcal{L}_{cyc}$         \\ \hline
                \multirow{2}{*}{\textbf{IHDP}($\epsilon_{PEHE}$)}   & in-sample   & $\textbf{.52} \pm \textbf{.0}$ & $.64\pm .1$ & $.59\pm .1$ & $.55\pm .0$  \\ \cline{2-6}
                                                        & out-sample   & $\textbf{.60}\pm \textbf{.1}$ & $.77\pm .1$ & $.75\pm .1$ & $.61\pm .1$  \\ \hline
                \multirow{2}{*}{\textbf{IHDP}($\epsilon_{ATE}$)}   & in-sample   & $\textbf{.10}\pm \textbf{.01}$ & $.11\pm .01$ & $.16\pm .03$ & $.12\pm .02$  \\ \cline{2-6}
                                                        & out-sample   & $\textbf{.13}\pm \textbf{.02}$ & $.20\pm .02$ & $.15\pm .02$ & $.15\pm .02$  \\  \hline
                                                        
                \multirow{2}{*}{\textbf{Jobs}(${R_{pol}}$)}   & in-sample   & $\textbf{.130}\pm \textbf{.0}$ & $.168\pm .0$ & $.167\pm .0$ & $.170\pm .0$  \\ \cline{2-6}
                                                        & out-sample   & $\textbf{.280}\pm \textbf{.0}$ & $.318\pm .0$ & $.317\pm .0$ & $.309\pm .0$  \\  \hline
                                                        
                \multirow{2}{*}{\textbf{Twins}({AUC})}   & in-sample   & $\textbf{.858}\pm \textbf{.001}$ & $.851\pm .001$ & $.856\pm .001$ & $.852\pm .001$  \\ \cline{2-6}
                                                        & out-sample   & $\textbf{.883}\pm \textbf{.001}$ & $.874\pm .001$ & $.881\pm .001$ & $.875\pm .001$  \\  \hline

\end{tabular}
\end{table*}



 Regression based methods directly model the covariables and the treatment, which suffer from high generalization error. Matching models consider similarity information, but they do not balance the distribution discrepancy well. By observing results, we find out that on each dataset, methods based on representation learning, such as CFR-Wass and SITE, are almost better than the methods based on regression adjustment and matching, such as OLS/LR$_1$ and k-NN. 
It proves the necessity and importance of eliminating distribution shift. Note that the distribution discrepancy damages the performance of counterfactual inference if we do not considering to reduce it.

Among all of methods based on representation learning, our model almost achieves best results on three datasets simultaneously. It proves the generalization and efficiency of our method in various domains. See that the SITE and ABCEI are better than BNN and CFR-Wass. We can conclude that considering information loss is helpful for counterfactual inference task. On the other hand, our method is better than SITE and ABCEI on almost every metric. It proves that the information loop by auto-encoders preserves original data properties and our method does the best at trading off between balancing distribution shift and preserving highly predictive information. Results on three real-world datasets demonstrate that CBRE has the best generalization performance.

\begin{figure*}[!htb]
\centering
\begin{subfigure}[t]{.3\textwidth}
  \centering
  \includegraphics[width=\textwidth]{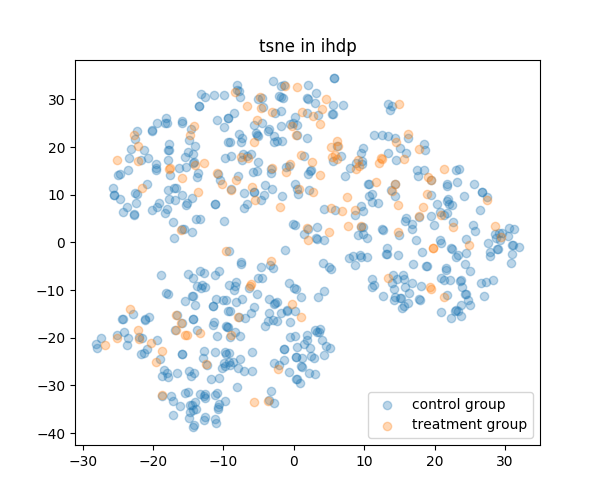}  
  \caption{Raw data of IHDP}
  \label{fig:a}
\end{subfigure}
\begin{subfigure}[t]{.3\textwidth}
  \centering
  \includegraphics[width=\textwidth]{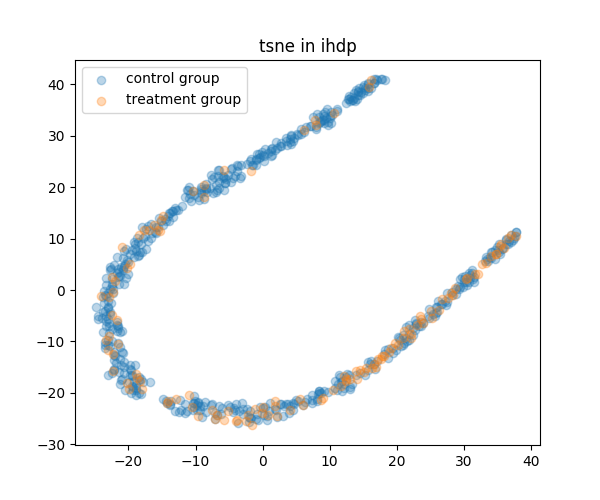}  
  \caption{Representations of CBRE on IHDP}
  \label{fig:b}
\end{subfigure}
\begin{subfigure}[t]{.3\textwidth}
  \centering
  \includegraphics[width=\textwidth]{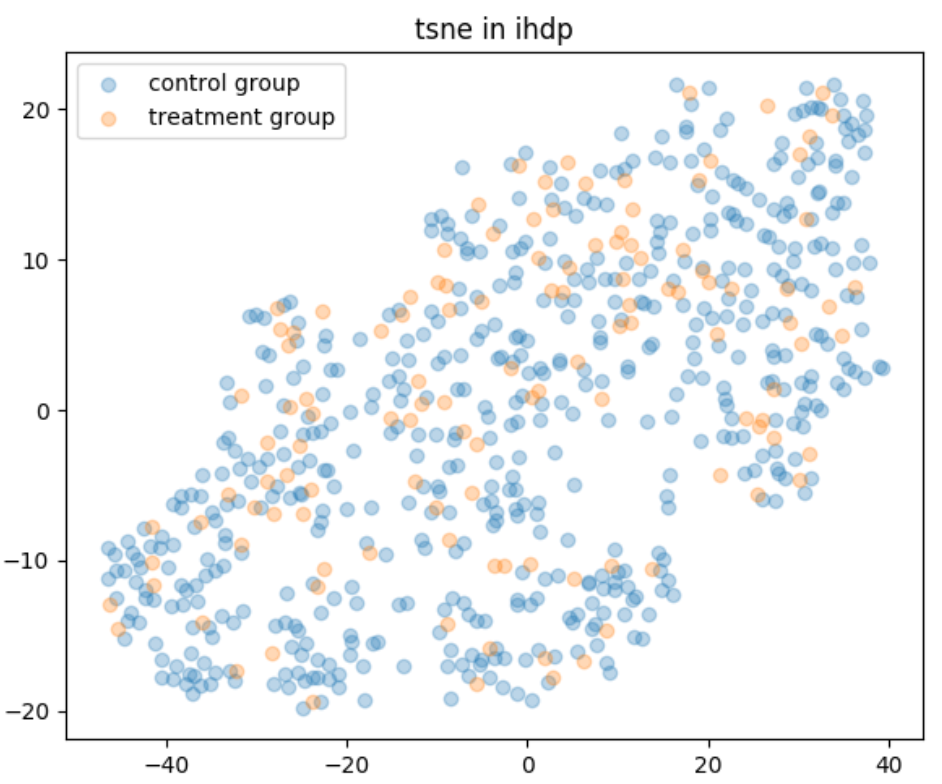}  
  \caption{Representations of SITE on IHDP}
  \label{fig:c}
\end{subfigure}


\begin{subfigure}[t]{.3\textwidth}
  \centering
  \includegraphics[width=\textwidth]{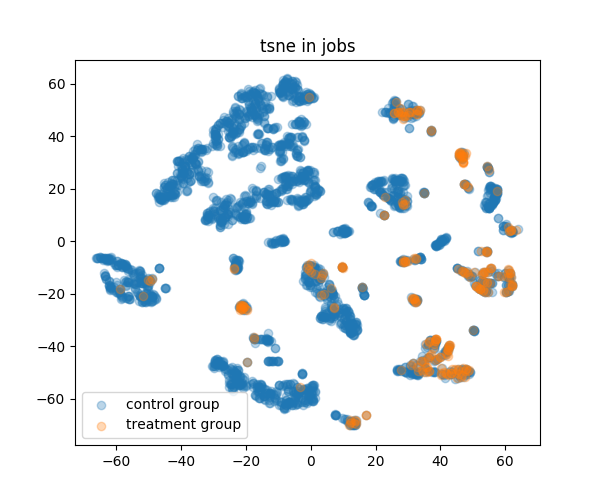}  
  \caption{Raw data of Jobs}
  \label{fig:d}
\end{subfigure}
\begin{subfigure}[t]{.3\textwidth}
  \centering
  \includegraphics[width=\textwidth]{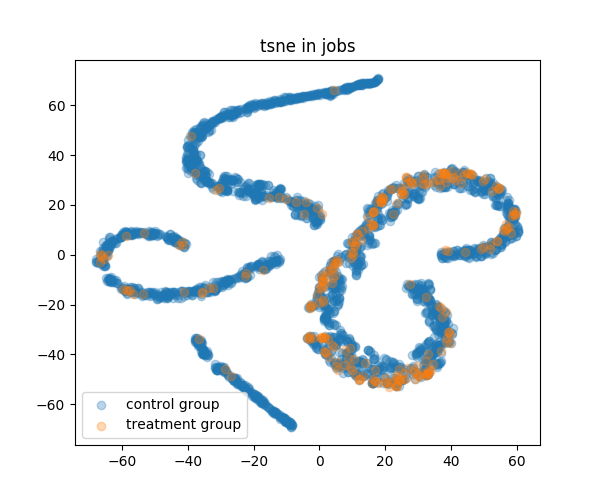}  
  \caption{Representations of CBRE on Jobs}
  \label{fig:e}
\end{subfigure}
\begin{subfigure}[t]{.3\textwidth}
  \centering
  \includegraphics[width=\textwidth]{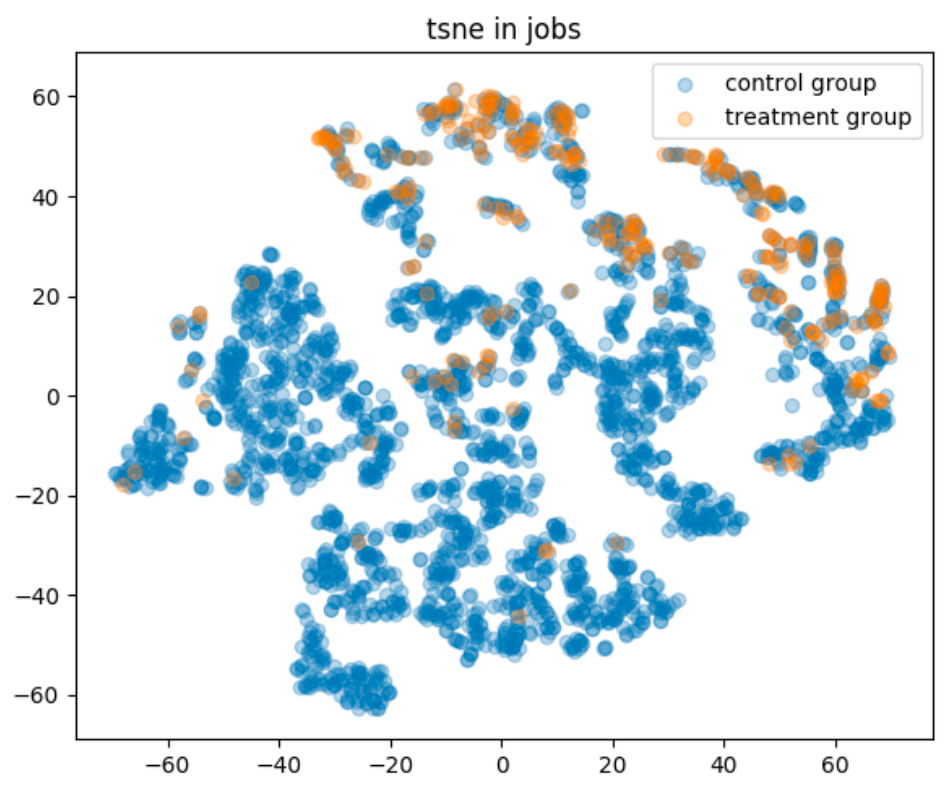}  
  \caption{Representations of SITE on Jobs}
  \label{fig:(f)}
\end{subfigure}


\caption{t-SNE visualization of treatment and control group, on IHDP and Jobs. The blue dots are control units and orange dots are treated units. The first row presents the raw data, representations learned by CBRE and representations of SITE method on IHDP dataset. The bottom figures refer to Jobs. Notice that our method realizes more balanced distributions between different groups in latent representation space.}
\label{fig:t-SNE}
\end{figure*}

It should be noted that we achieve $30\%$ improvement in terms of  $\sqrt{\epsilon_{PEHE}}$ on IHDP dataset in both within-sample and out-of-sample cases. Overall, by designing the discriminator and two decoders, CBRE is able to efficiently preserve overall highly predictive information and meanwhile drop bias information to balance the distributions of different groups.  


\subsection{Ablation Study}
We perform the ablation experiment to examine contributions of each component in CBRE on final inference performance. We separately test the improvement of the adversarial training and the information loop to the final effect. The results are shown in Table \ref{table_Ablation Study}.  
Firstly, we keep only the encoder and prediction network in Figure ~\ref{fig:CBRE_model}. The objective loss is reduced to $\mathcal{L}_p$ plus regularization term.     

Then, on the basis of the first, we add the discriminator to evaluate its contribution for final performance. So total loss will accumulate $\mathcal{L}_D$. As we see in Table \ref{table_Ablation Study}, the discriminator is helpful to improve performance by 1\% to 25\% on different metrics. It corresponds to that the adversarial training adjusts the latent representations positively.

At last, we evaluate the role of the information loop. The loss is $\mathcal{L}=\mathcal{L}_p+\mathcal{L}_{rec}+\mathcal{L}_{cyc}$. It can be observed that the model achieves better performance if it adds decoder modules that build the information loop. We also compare it with $\mathcal{L}_p+\mathcal{L}_d$. It is worse than the latter on $R_{pol}$ of Jobs and AUC of Twins. Considering the Jobs and Twins are larger than IHDP dataset, we guess that, in this case, distribution shift has more serious damage to the final effect when the dataset size is large. Therefore, balancing distribution is also important for the information loop.

We combine all of above modules, and our model achieves the best performance.

\subsection{Latent Representation Visualization}
We conduct the t-SNE visualization  \cite{van2008visualizing} of original data and corresponding latent representations on three datasets, as Figure \ref{fig:t-SNE} shows. To prove the effectiveness for distribution shift, we separate both raw data and representations of the treatment and control group. It is obvious on Jobs that distribution discrepancy exists and representations are well merged after training. The distribution discrepancy can be noticeable on IHDP, and corresponding reductions in discrepancy are also be noticed. 
We find out that the overlap between the treatment group and control group in latent representation space is obvious. This denotes the effectiveness of adversarial training in balancing distribution discrepancy. And meanwhile the information loop further reduces the distribution shift, and they both ensure the model performance.  

We selectively add the representations of SITE on IHDP and Jobs due to space  restriction. SITE proposes to balance distribution shift and preserve local similarity. As the Figure \ref{fig:t-SNE} shows, compared to raw data, latent representations of SITE indeed demonstrate the reduction in distribution shift. However, the discrepancy between the treatment and control group is also obvious. It is clear that our method achieves more balanced distributions in latent representation space due to preserving global data properties.

\subsection{Hyper-parameter Optimization}

Due to the fact of missing counterfactual outcomes, standard cross-validation methods are not able to perform. We follow the procedure in  \cite{shalit2017estimating} to optimize hyper parameters.  

In particular, we search the learning rate in $\{1e$-$2$, $1e$-$3$,$1e$-$4\}$, the depth of each module in $\{3,4,5,6\}$, the dimension of each module in $\{50, 100, 200, 300 \}$, the batch size from 50 to 200 with increments of 10, $\{ \alpha$, $\beta$, $\gamma\}$ from 0.5 to 1.5 with increments of 0.1. The optimal hyper-parameters for three datasets can be found in the supplementary material. We set parameters of baseline models as same as what in the original papers.

\section{Related Works}
\label{related works}

Recently, counterfactual inference attracts considerable attention in many fields. In this section, we present existing works related to counterfactual inference. We divide approaches in this area into three categories.

{\bf Matching methods.} Matching methods estimate effects by finding the nearby samples. Specifically, the counterfactual outcome of a sample to a treatment is similar to the observed factual outcome of its nearest neighbours that receive the same treatment \cite{crump2008nonparametric}.  

{\textbf{Regression adjustment.}} Regression based models fit a supervised learning model on the features and the treatment to estimate the potential outcomes, such as OLS/LR$_1$ and OLS/LR$_2$. An advanced tree-based machine learning method, like BART \cite{chipman2010bart}, is also used. 

{\textbf{Representation learning.}} Johansson et al. \cite{johansson2016learning} utilize a multi-layer neural network to obtain representations of covariables. And they concatenate representations with treatment assignment variables to feed a prediction network. Shalit et al. \cite{shalit2017estimating} build on and extend work by  Johansson. The proposed framework is end-to-end and is able to learn non-linear representations by integrating Integral Probability Metrics. To keep influence of the treatment on prediction network, they use two separate networks to model predictions of different treatments. Yao et al. \cite{yao2018representation} consider not only distribution bias but also local similarity information and propose their method SITE. SITE maps triple sample pairs into the latent space, and they propose position-dependent deep metric and middle-point distance minimization to constrain the representation process. 
However, only triple samples are used, and it limits the performance to only consider local similarities. And meanwhile, hard samples are selected by the propensity score model that is not robust to misspecification. It's unavoidable to trade off between preserving highly predictive information and reducing distribution shift. Our work is perhaps most similar to ABCEI \cite{du2021adversarial} because we both consider preserve data properties when transforming into latent space from a global perspective. \cite{du2021adversarial} employs mutual information metric to reduce information loss. However, mutual information in high-dimension space is hard to compute. And it is not accurate enough to take concatenation of shuffled covariates and representations as product of marginals. Practical performance also validates that our method is better. 

Matching and regression methods do not take distribution discrepancy into consideration, which damages the performance for counterfactual inference. Inherent problem still remains unsolved in methods based on representation learning. It is unavoidable that information loss exists when transforming raw data into latent space.  

Motivated by advances of supervised learning and domain adaptation \cite{hoffman2018cycada, chen2019distributionally, 10.1145/3108257}, we introduce an information loop to preserve information, and we realize a better estimator for individual treatment effect (ITE). In the experiment section, we show that our method matches/outperforms the baselines.




\section{Conclusion}

We propose a novel framework based on cycle-balanced representation learning for counterfactual inference. By aid of adversarial training and information loop, we reduce distribution discrepancy and preserve original data properties to estimate individual treatment effects. We perform extensive experiments on three real-world datasets and compare our method with multiple state-of-the-art models.
The results on the IHDP, Jobs, and Twins datasets demonstrate that our method outperforms the baselines in almost each case. It proves the effectiveness of our proposed model, and great generalization ability on different datasets from various domains. Extensive evaluations and experiments validate the advantages of reducing distribution shift and building information loop in the task of counterfactual inference. 

We focus on binary treatment assignments in this paper, which is common in practice. Maybe we only need to add a few decoders by aid of clustering assignments so that our framework can be scaled to multiple treatments. We leave this to future work. 
We also look forward to relaxing unconfoundedness assumption and consider finding hidden confounders by data-driven methods, and incorporating this point with our method to formulate a novel framework.

\end{document}